\documentclass[journal,web]{article}
\usepackage{authblk}
\usepackage{framed,multirow}
\usepackage{amssymb}
\usepackage{latexsym}
\usepackage{amsmath}
\usepackage{subfig}
\usepackage{floatflt}
\usepackage{url}
\usepackage{xcolor}
\usepackage{hyperref}
\usepackage{graphicx}

\providecommand{\keywords}[1]{\textbf{\textit{\\Keywords---}} #1}
\date{}

\renewcommand\footnotemark{}

\title{From Generalization to Precision: Exploring SAM for Tool Segmentation in Surgical Environments}

\author{Kanyifeechukwu J. Oguine, Roger D. Soberanis-Muku, Nathan Drenkow, Mathias Unberath}

\begin{document}
\maketitle

\begin{abstract}
\textbf{Purpose:}
Accurate tool segmentation is essential in computer-aided procedures. However, this task conveys challenges due to artifacts' presence and the limited training data in medical scenarios. Methods that generalize to unseen data represent an interesting venue, where zero-shot segmentation presents an option to account for data limitation. Initial exploratory works with the Segment Anything Model (SAM) show that bounding-box-based prompting presents notable zero-short generalization. However, point-based prompting leads to a degraded performance that further deteriorates under image corruption. We argue that SAM drastically over-segment images with high corruption levels, resulting in degraded performance when only a single segmentation mask is considered, while the combination of the masks overlapping the object of interest generates an accurate prediction.
\textbf{Method:}
We use SAM to generate the over-segmented prediction of endoscopic frames. Then, we employ the ground-truth tool mask to analyze the results of SAM when the best single mask is selected as prediction and when all the individual masks overlapping the object of interest are combined to obtain the final predicted mask. We analyze the Endovis18 and Endovis17 instrument segmentation datasets using synthetic corruptions of various strengths and an In-House dataset featuring counterfactually created real-world corruptions.
\textbf{Results:}
Combining the over-segmented masks contributes to improvements in the IoU. Furthermore, selecting the best single segmentation presents a competitive IoU score for clean images.
\textbf{Conclusions:} 
Combined SAM predictions present improved results and robustness up to a certain corruption level. However, appropriate prompting strategies are fundamental for implementing these models in the medical domain.
\end{abstract}


\section{Introduction}
\label{sec:intro}
The automatic segmentation of instruments plays a crucial role in computer-assisted interventions, where applications such as surgical understanding, tool tracking, tool-tissue iteration analysis, and navigation can benefit from tool segmentation models. These applications require precise and robust tool segmentation algorithms capable of working under the challenging environment that surgical images impose, where  smoke, motion blur, and brightness and intensity variations can be present during the procedure~\cite{bib:ali2020}. Over the years, numerous image segmentation models have been developed, gradually progressing from conventional methods like clustering, edge-based, and contour-based detection towards deep learning models such as U-Net~\cite{Ronneberger2015Nov} and SegNet~\cite{Badrinarayanan2017Jan}. Deep learning models have demonstrated exceptional performance in segmentation tasks, yet they require large datasets for good generalization and robustness to corruptions in medical imaging. The availability of adequate training medical datasets presents a limitation in the medical imaging analysis domain. 

Recently, advancements in foundational models like Large Language Models (LLMs)~\cite{Devlin2019Jun,OpenAI2023Mar}, and recently in the image domain with the Segment Anything Model (SAM)~\cite{Kirillov2023Apr} have revolutionized our perception of model generalization. SAM has established a notable reputation for its exceptional performance and abstraction across diverse unseen datasets, leading to a promising avenue toward efficient surgical tool segmentation. However, it is known that the performance of segmentation algorithms can be compromised by factors commonly encountered in surgical scenarios, such as variations in lighting, environmental and acquisition conditions, and imaging hardware. While SAM has demonstrated exceptional performance in general real-world image segmentation, its applicability to surgical environments still needs to be explored.

Initial exploratory works with SAM~\cite{Wang2023Apr} have shown that bounding-box-based prompting presents notable zero-short generalization. However, point-based prompting leads to a degraded performance that further deteriorates under image noise and corruption. In contrast, we argue that SAM can generate an accurate mask provided that all individual regions overlapping the object of interest are consolidated into a single mask. This can suggest that, while SAM drastically over-segment images with high corruption levels, resulting in degraded performance when only a single segmentation mask is considered, it still manages to accurately detect object boundaries while hallucinating additional ones. 
Under this perspective, this work evaluates the stability of SAM in tool segmentation under different levels of simulated and real image corruptions. We consider the scenario when the prediction is composed of a single segmentation mask and compare it with the case when predictions are composed of multiple sub-masks that overlap the region of interest (ROI).

\begin{figure}[t]
\centerline{\includegraphics[width=0.9\textwidth]{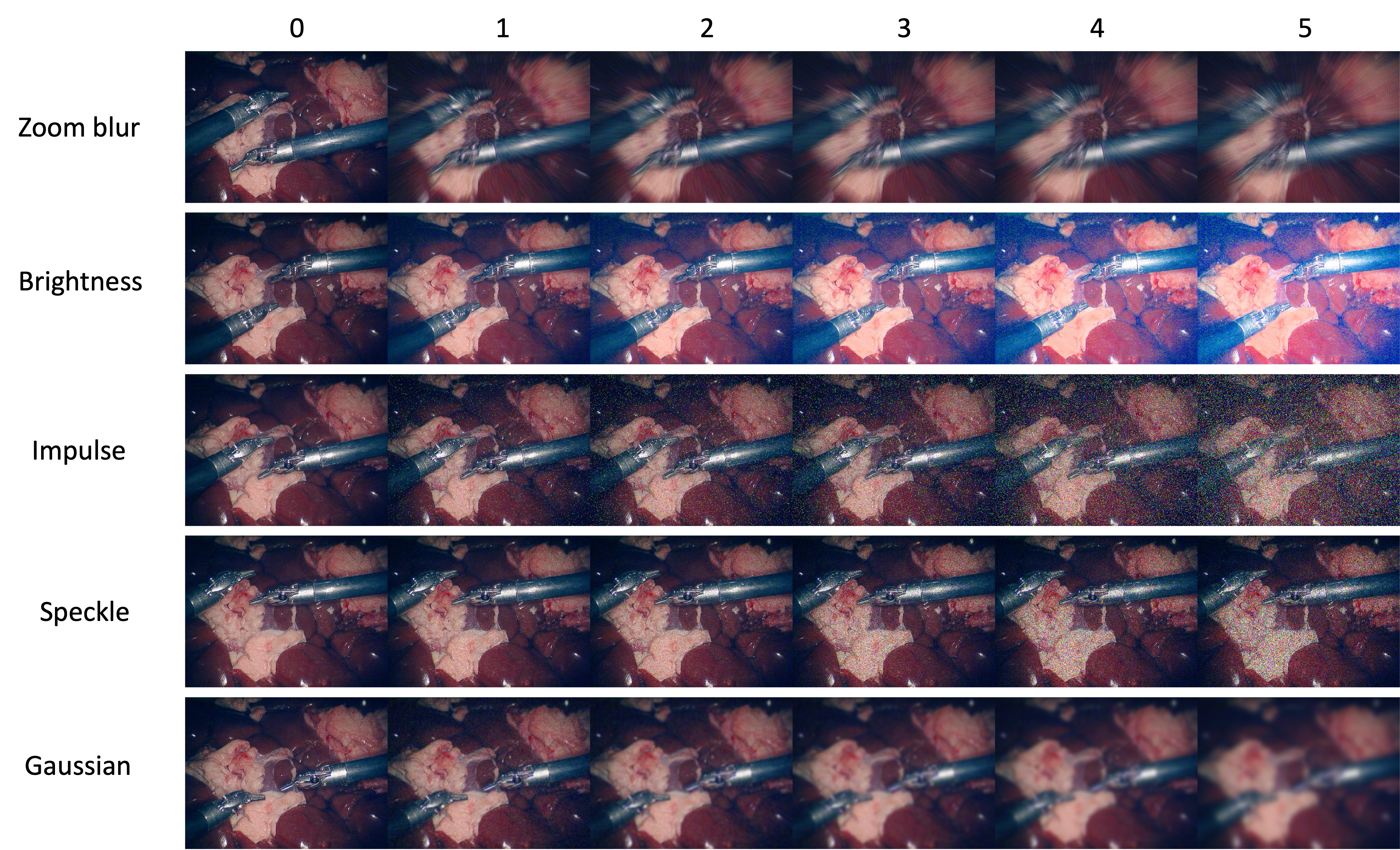}}
    \caption{Examples of five types of corruptions applied to the images with their corresponding severity level. Images from our In-House dataset.}
    \label{fig:noise_samples}
\end{figure}

\section{Method}

To perform the evaluation, we employ SAM's automatic segmentation with default parameters to generate a grid-based over-segmented version of the image. 
We select the predictions that overlap the ROI defined by the tools' ground-truth mask to generate the tools' segmentation from the complete set of sub-masks. A sub-mask overlaps the ROI if its intersection ratio ($i_r$) with the ROI is more than 50\,\%. This ratio is defined by Eq.~\ref{eq:rel_area}, where $M_{ROI}(x)$ represents the binary ground-truth mask, $M_{s}(x)$ a particular binary segmentation sub-mask obtained with SAM, and $x$ is a pixel position.
\begin{equation} 
\label{eq:rel_area}
     i_r = \frac{\sum_x M_{ROI}(x) \cdot M_s(x)}{\sum_x M_s(x)}
\end{equation}
This will lead to a set $S = M_1, M_2, \dots, M_n$ of $n$ overlapping sub-masks. 
The predicted segmentation is generated considering two scenarios. The first scenario evaluates the robustness and stability of the best single sub-mask as defined by Eq. \ref{eq:best_single}. 
\begin{equation}
\label{eq:best_single}
M_{single} = \arg \max_i \sum_x M_i(x), \text{ } M_i \in S, \text{ } i = 1, 2, \dots, n
\end{equation}
The second scenario combines all the overlapping sub-masks to generate the final prediction, using Eq. \ref{eq:combined_mask} where ``$+$" defines the pixel-wise addition between the binary masks.  

\begin{equation}
\label{eq:combined_mask}
M_{comb} =  M_1 + M_2 + \cdots + M_{n}
\end{equation}

Overall, the single mask segmentation can be interpreted as the segmentation obtained when a point-based prompt is placed in a representative section of the tool. Using a similar analogy, we can understand the combined segmentation resulting from placing multiple prompts in the tools. 
\begin{figure}[t]
    \centerline{\includegraphics[width=0.99\textwidth]{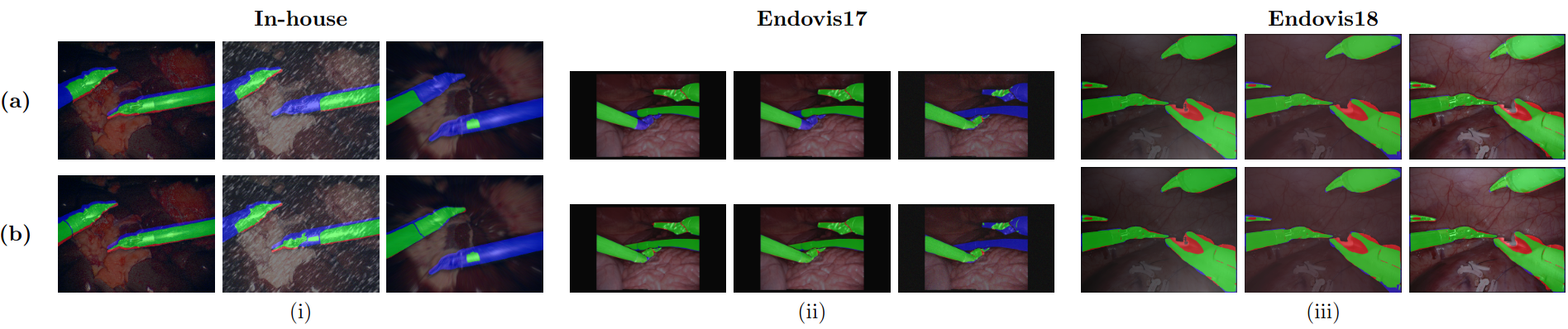}}
    \caption{Overlay of (a) single and (b) combine prediction masks for different corruption types. True positives, false positives, and false negatives are indicated in green, red, and blue, respectively. Samples of the three datasets are presented.}
    \label{fig:samples}
\end{figure}

We evaluate the performance of SAM on three endoscopic datasets, including EndoVis17~\cite{Allan2020Jan}, EndoVis18~\cite{Allan2019Feb}, and an In-House dataset. Both EndoVis datasets contain binocular frames. In our experiments, we use the left frames of the EndoVis17 and EndoVis18 training split, leading to 225 and 149 images for each dataset, respectively. Our In-House stereo endoscopic dataset~\cite{Ding2022Sep} was collected using the da Vinci Research Kit (dVRK)~\cite{Kazanzides2014May} and includes 400 frames. Additionally, the In-House contains three counterfactually created real-world corruptions. Note that the In-House dataset does not include ground-truth tool masks. We generated the annotations using a semi-automatic approach, employing  SAM to generate the over-segmented image and then manually correcting the segmentation masks. 
We apply 18 image corruptions to the images of each with five levels of severity. Fig. \ref{fig:noise_samples} present examples for five of these corruptions for reference. We then obtain the single and combined segmentation considering corruption/severity groups. We report the average intersection over union (IoU) per group in our experiments. Note that to test the zero-shot generalizability of the network, the segment anything model was used out-of-the-box with its predefined weights, with no additional fine-tuning to the datasets.

\section{Results}

\begin{figure}[h]
  \centering
  \subfloat{\includegraphics[width=0.5\textwidth]{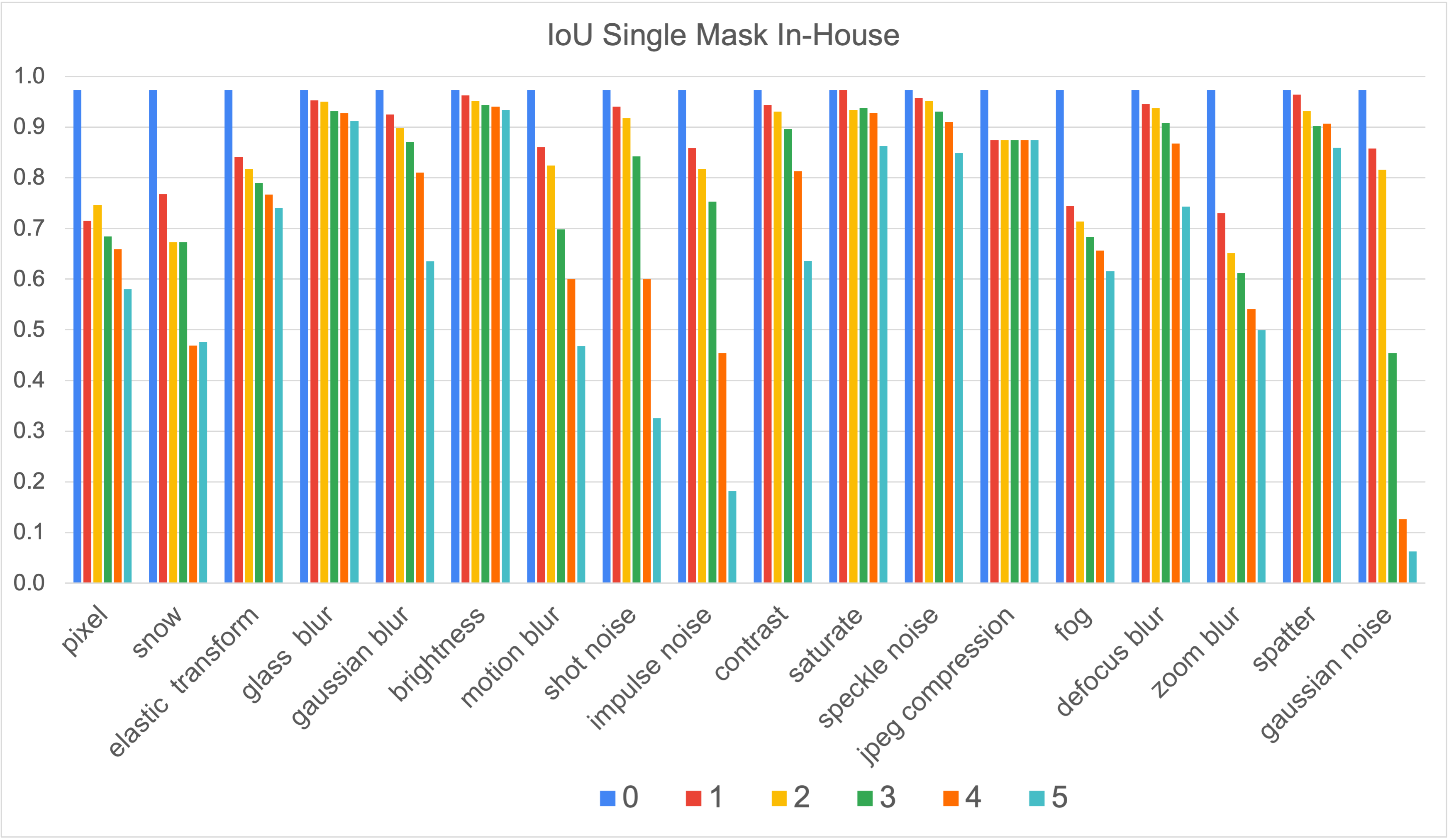}}
  \subfloat{\includegraphics[width=0.5\textwidth]{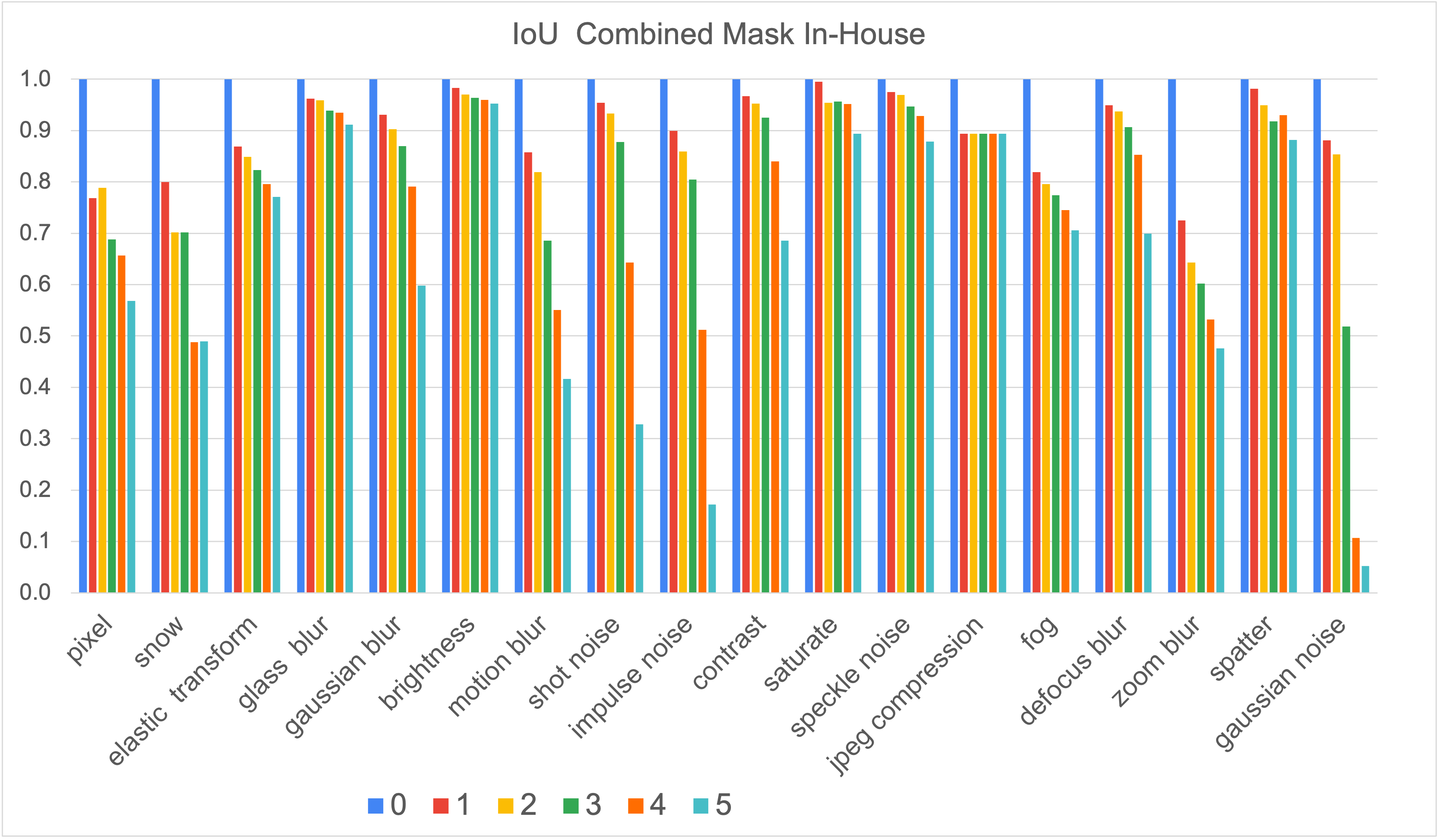}}
  \caption{Average IoU for the 18 types of corruption, with five levels of severity for the Single v.s. Combined SAM segmentation mask in the In-House dataset.}
  \label{fig:in-house}
\end{figure}

\begin{figure}[h]
  \centering
  \subfloat{\includegraphics[width=0.5\textwidth]{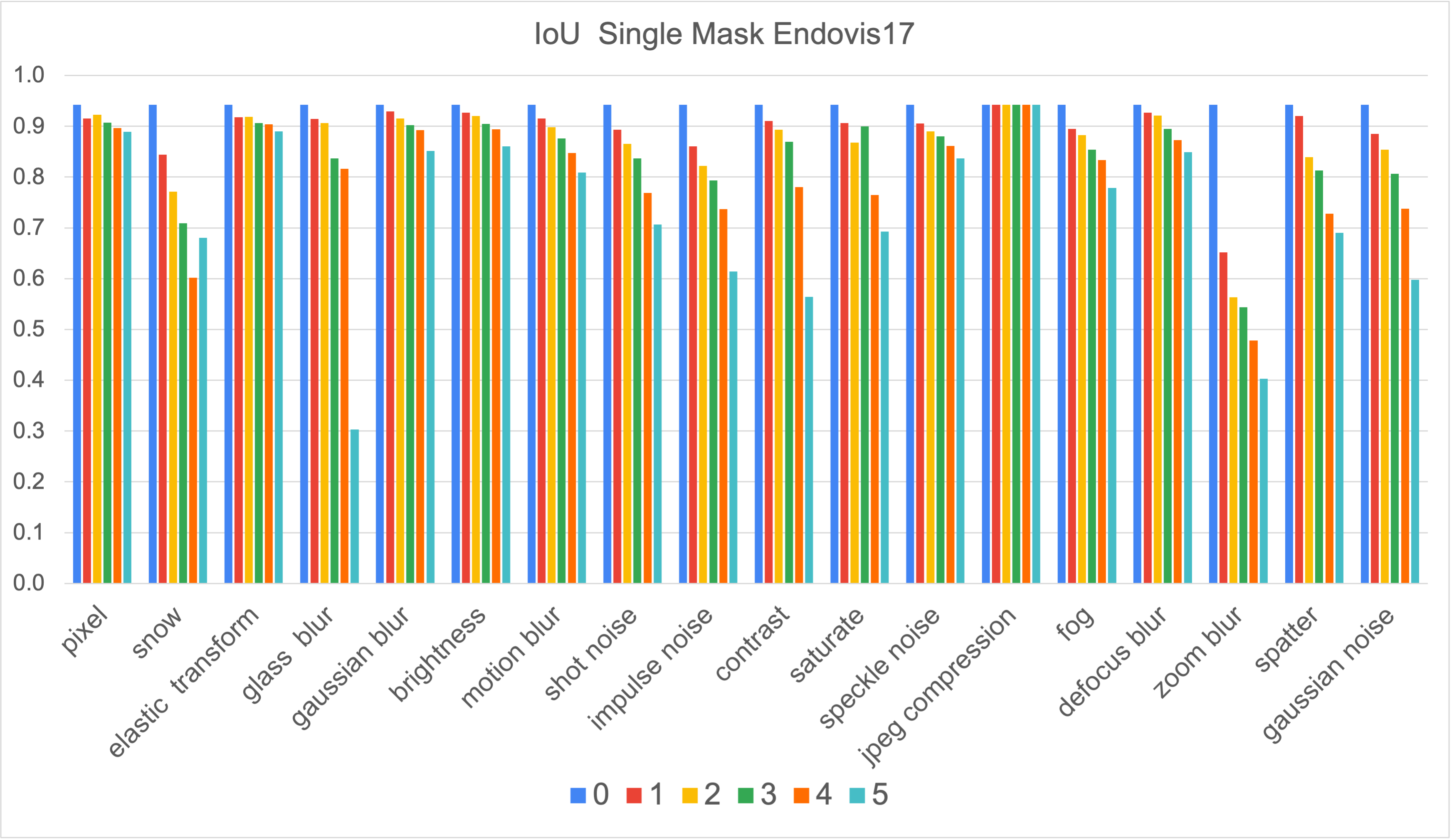}}
  \subfloat{\includegraphics[width=0.5\textwidth]{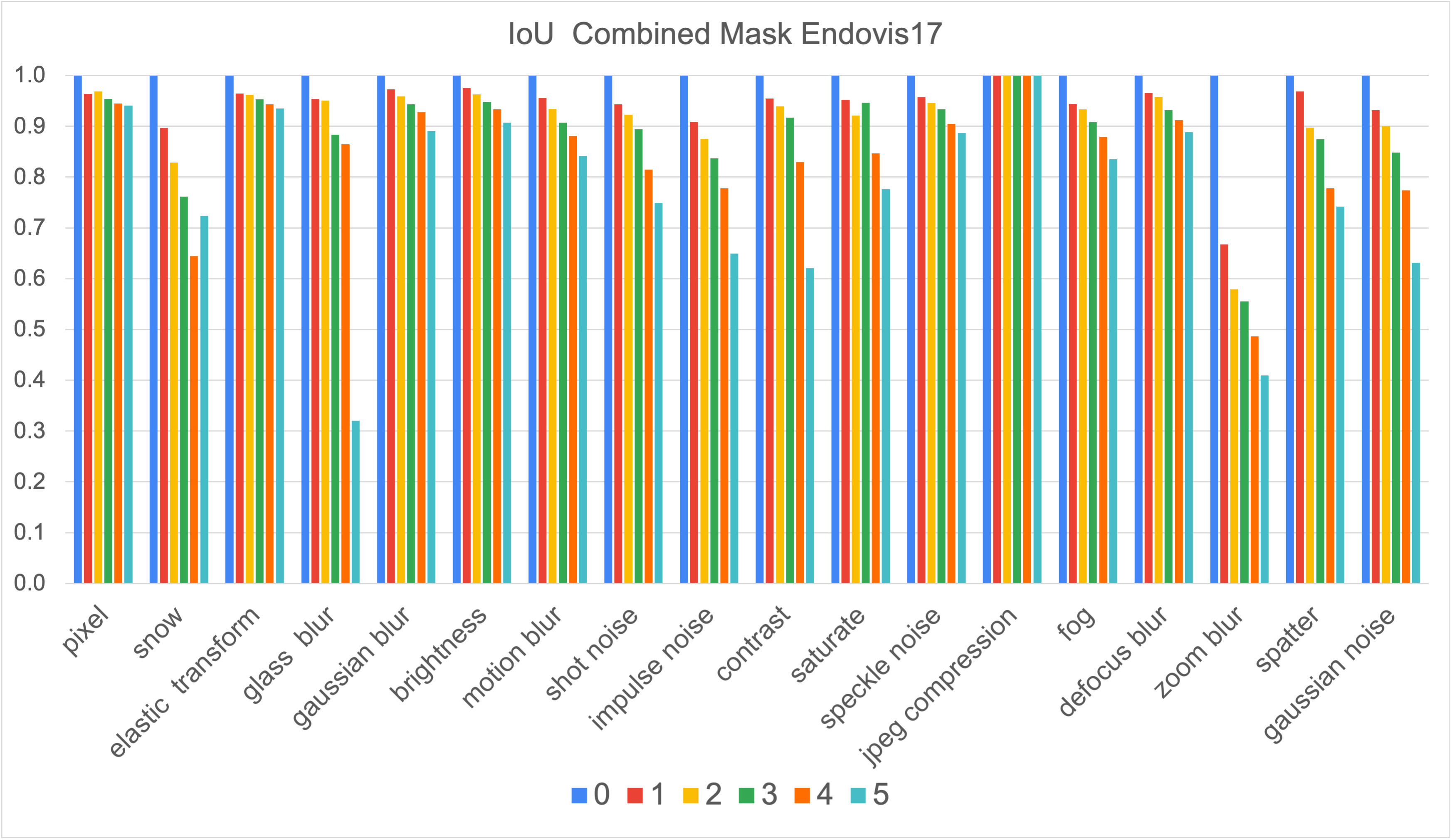}}
  \caption{Average IoU for the 18 types of corruption, with five levels of severity for the Single v.s. Combined SAM segmentation mask in the EndoVis17 dataset.}
\label{fig:endovis17}
\end{figure}

\begin{figure}[h]
  \centering
  \includegraphics[width=0.49\textwidth]{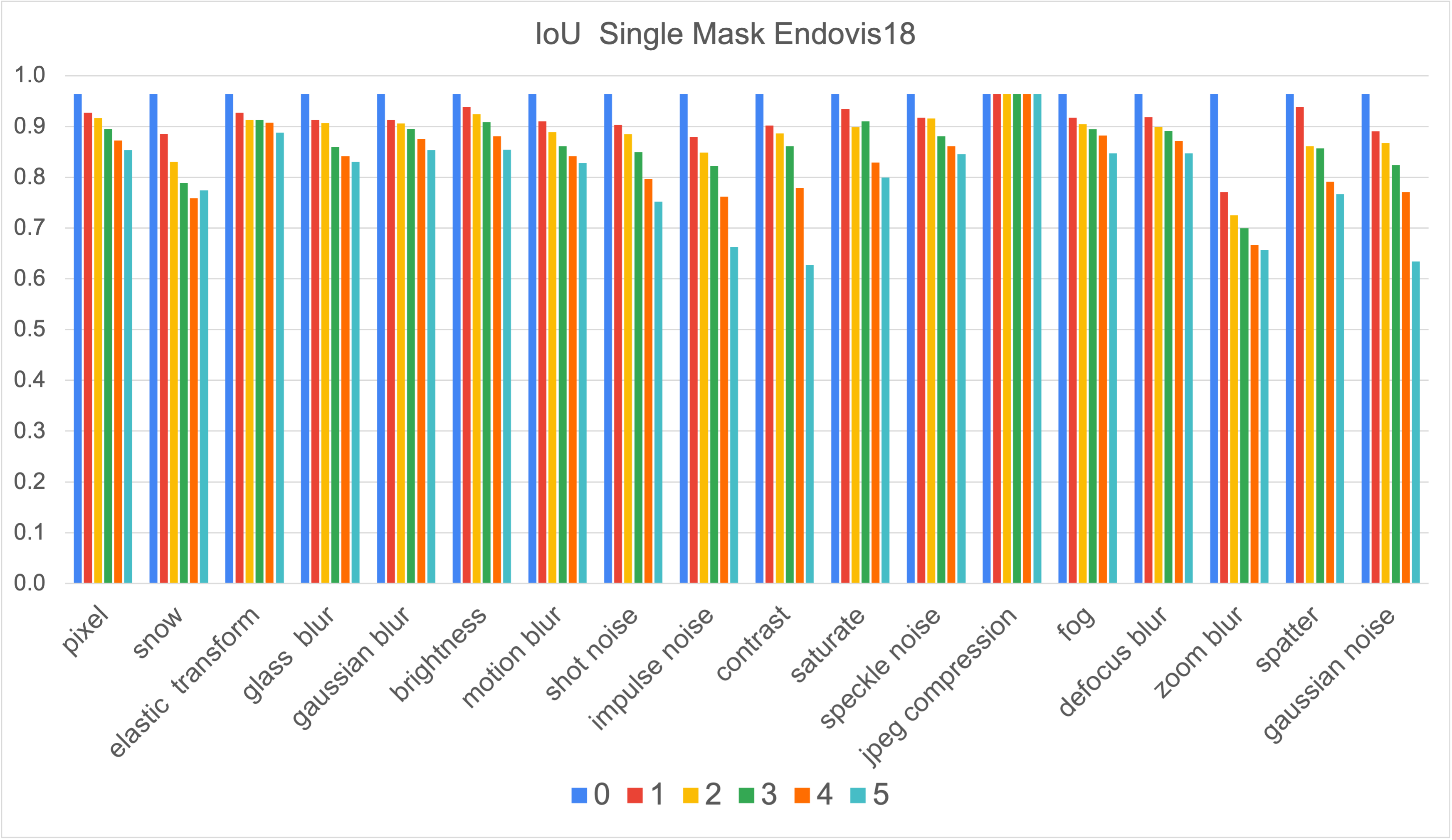}
  \includegraphics[width=0.49\textwidth]{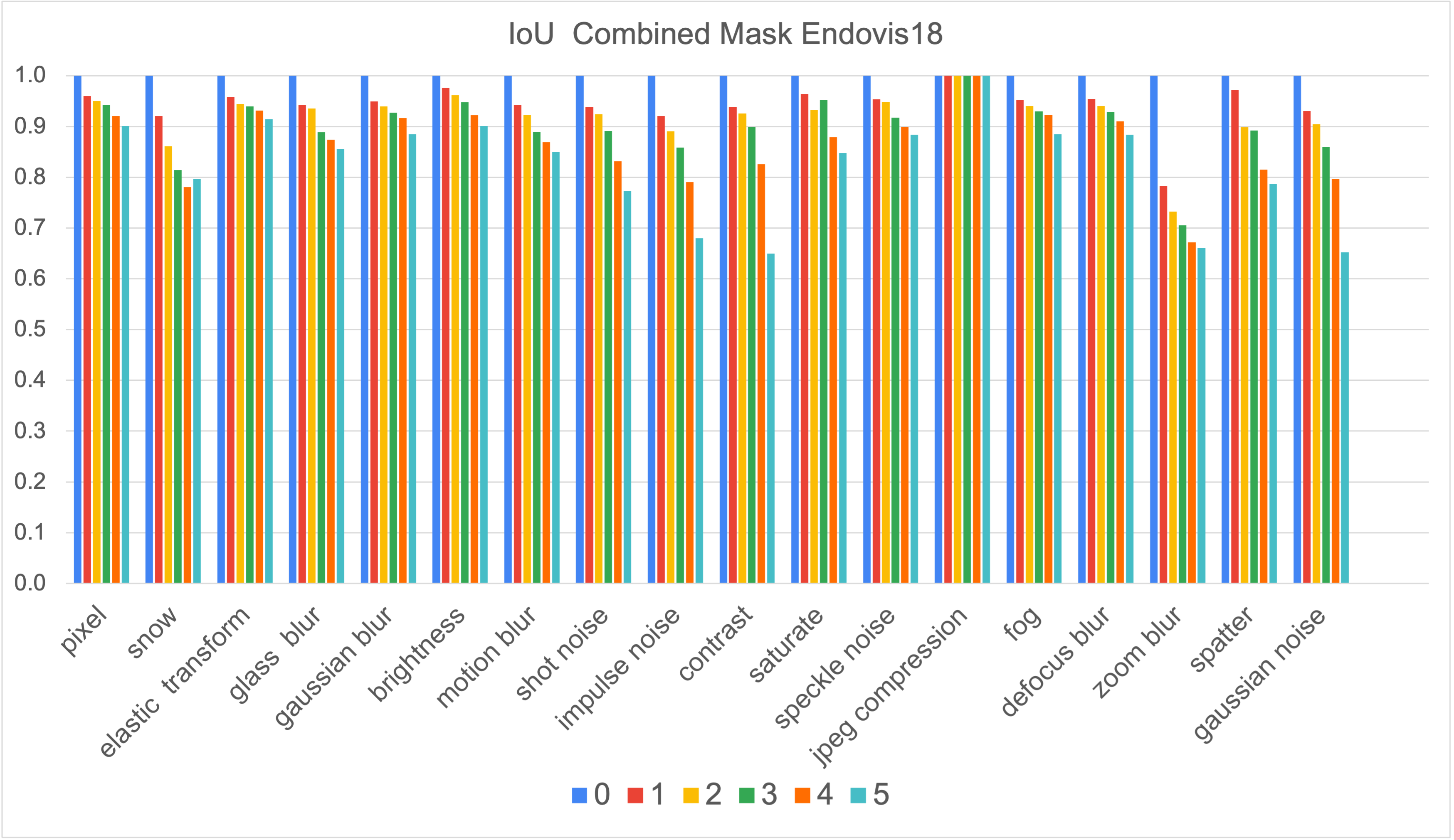}
  \caption{Average IoU for the 18 types of corruption, with five levels of severity for the Single v.s. Combined SAM segmentation mask in the EndoVis18 dataset.}
\label{fig:endovis18}
\end{figure}

\begin{figure}[h]
  \centering
  \includegraphics[width=0.9\textwidth]{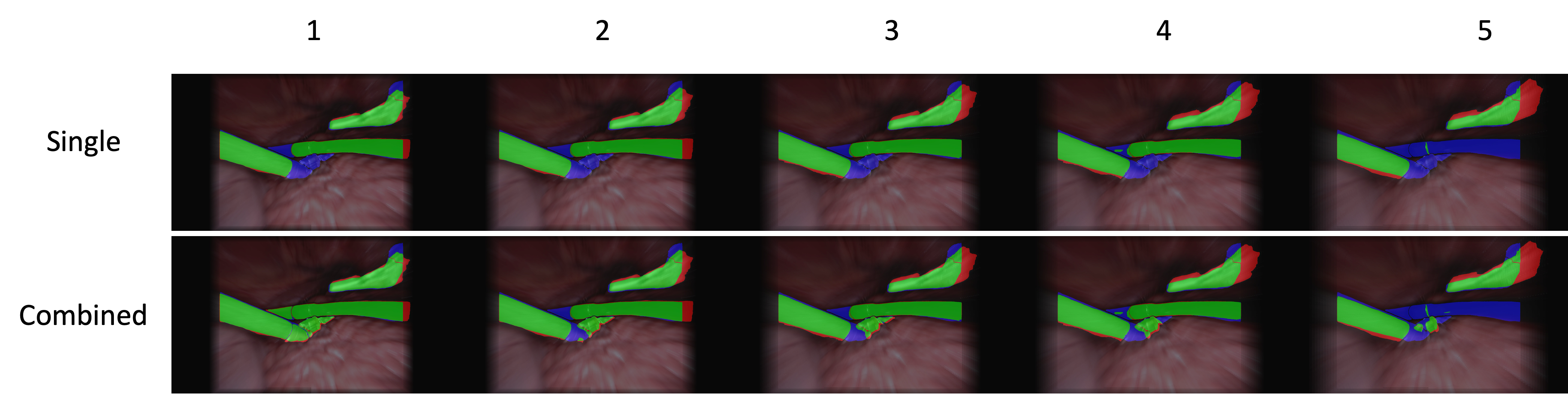}  
  \caption{Frame segmentation visual performance in the presence of different levels of zoom blur. The figure presents the segmentation of the single and combined masks. True positives, false positives, and false negatives are indicated in green, red, and blue, respectively.}
\label{fig:zoom}
\end{figure}

\begin{figure}[h]
  \centering
  \includegraphics[width=0.9\textwidth]{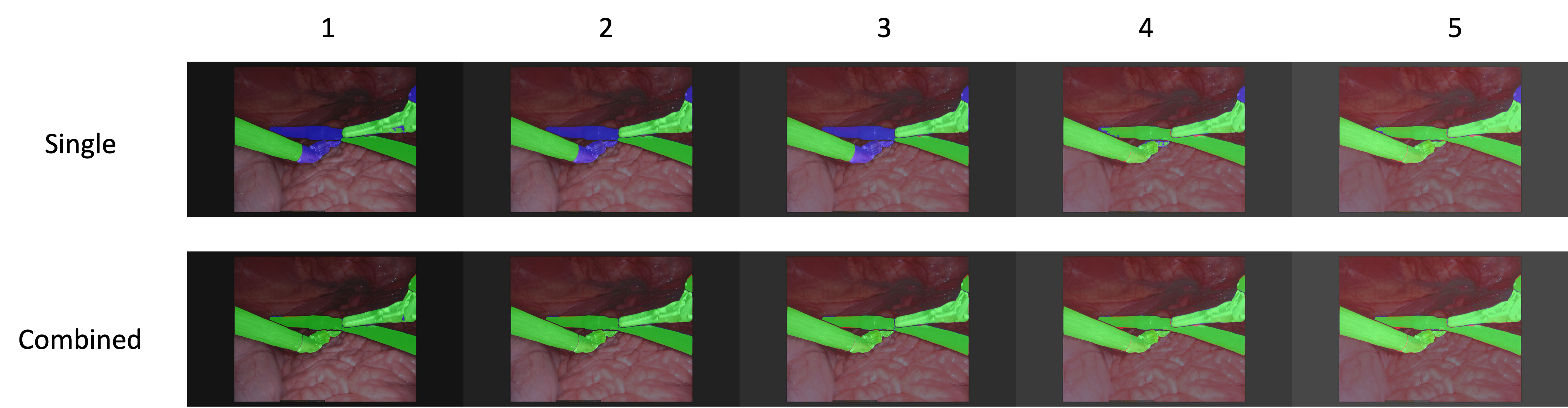}  
  \caption{Frame segmentation visual performance in the presence of different levels of brightness. The figure presents the segmentation of the single and combined masks. True positives, false positives, and false negatives are indicated in green, red, and blue, respectively.}
\label{fig:brightness}
\end{figure}

Results for the single (s) and combined (c) mask are presented in Figs. \ref{fig:in-house}, \ref{fig:endovis17}, and \ref{fig:endovis18} for the In-House, EndoVis17, and EndoVis18 datasets, respectively. Similarly, the performance of the segmentation masks in the real perturbations is presented in Table \ref{tab:in-house-real}. Additionally, Fig. \ref{fig:samples} shows visual examples of segmented images in the different datasets. Results show that the overall performance of the SAM degrades with the severity of the corruption but at different levels, depending on the type of perturbation. For example, it is suggested that zoom blur (Fig. \ref{fig:zoom}) causes a high degradation in the performance consistently across the datasets and predicted mask, even at the first level of severity, in contrast to brightness (Fig. \ref{fig:brightness}) or JPEG compression, where the performance is more stable for both single and combined masks. 

When comparing the outcomes of the single prediction with the combined mask, we can observe a relative improvement in the IoU when employing the combined segmentations, especially in both EndoVis datasets. This suggests combining the combined masks captures additional details missed by the single mask prediction. For example, it can be noticed in Fig. \ref{fig:zoom} and Fig. \ref{fig:brightness} that the single mask fails to segment the tooltip in different cases, while by using a combined mask, it is possible to obtain these details of the tool.

\begin{table}[h]
\centering
\caption{Average IoU for the three counterfactually generated corruption for the Single (s) v.s. Combined (c) SAM segmentation mask in the In-House dataset.}
\label{tab:in-house-real}
\resizebox{0.35\textwidth}{!}{
\begin{tabular}{ccccc}
\hline
level  & Mask & High & Low  & Smoke\\
&      & Brightness & Brightness &   \\
\hline
\multirow{2}{*}{0} & s & 0.97 & 0.97 & 0.97\\
 & c & 0.99 & 0.99 & 0.99\\
\hline
\multirow{2}{*}{1} & s	& 0.94 & 0.61 & 0.72\\
& c & 0.95 & 0.67 & 0.75\\
\hline
\end{tabular}%
}
\end{table}

Regarding stability to the presence of perturbations, both masks present a degree of robustness to the severity of the disturbances. However, this is more evident when evaluating the combined segmentation. In some cases, both masks present a high level of degradation in their performance. It is possible that, up to a specific type and level, the perturbations are strong enough to heavily modify the scene's visibility, leading to loss of information and the degradation of the results. For example, blur can reduce the contrast between the tool contour and the background, creating incorrect segmentations that contain a large portion of the tissue. However, it can be noticed in Fig. \ref{fig:zoom} that the use of a combined mask can reduce the impact of the noise and still recover sections of the tooltip.

\section{Conclusions}
The performance of SAM in the tool segmentation tasks can be affected by prompt selection. While a single mask can obtain adequate generalization and performance in this task, a prediction composed of multiple sub-masks can recover additional details of the tool, as suggested by an improvement in the IoU. This can indicate a tendency of SAM to over-segment the objects of interest and the necessity of defining adequate prompting strategies that lead to optimal results.  

\bibliographystyle{acm}
\bibliography{report} 

\end{document}